\documentclass[journal]{IEEEtran}
\pdfoutput=1
\usepackage{amsmath,amsfonts}
\usepackage[ruled,linesnumbered, longend, vlined]{algorithm2e}
\usepackage{array}
\usepackage{arydshln}
\usepackage[caption=false,font={rm,small},labelfont=rm,textfont=rm]{subfig}
\usepackage{textcomp}
\usepackage{stfloats}
\usepackage[hyphens]{url}
\usepackage{breakurl}
\usepackage{url}
\usepackage{verbatim}
\usepackage{graphicx}
\usepackage{cite}
\usepackage{url}
\usepackage{float}
\usepackage{color}
\usepackage{xcolor}
\usepackage{multirow}
\usepackage{tabularx}
\usepackage[para,online,flushleft]{threeparttable}
\usepackage{makecell}
\usepackage{tablefootnote}
\usepackage[para,online,flushleft]{threeparttable}
\usepackage{hyperref}
\usepackage[inline]{enumitem}
\usepackage{amssymb}
\usepackage{booktabs}
\usepackage[hang,flushmargin]{footmisc}

\begin{document}

\title{Large Language Model-Empowered Interactive Load Forecasting}

\author{Yu Zuo, Dalin Qin, and Yi Wang
}

\markboth{Preprint}%
{Shell \MakeLowercase{\textit{et al.}}: A Sample Article Using IEEEtran.cls for IEEE Journals}

\maketitle

\begin{abstract}
The growing complexity of power systems has made accurate load forecasting more important than ever. An increasing number of advanced load forecasting methods have been developed. However, the static design of current methods offers no mechanism for human–model interaction. As the primary users of forecasting models, system operators often find it difficult to understand and apply these advanced models, which typically requires expertise in artificial intelligence (AI). This also prevents them from incorporating their experience and real-world contextual understanding into the forecasting process. Recent breakthroughs in large language models (LLMs) offer a new opportunity to address this issue. By leveraging their natural language understanding and reasoning capabilities, we propose an LLM-based multi-agent collaboration framework to bridge the gap between human operators and forecasting models. A set of specialized agents is designed to perform different tasks in the forecasting workflow and collaborate via a dedicated communication mechanism. This framework embeds interactive mechanisms throughout the load forecasting pipeline, reducing the technical threshold for non-expert users and enabling the integration of human experience. Our experiments demonstrate that the interactive load forecasting accuracy can be significantly improved when users provide proper insight in key stages. Our cost analysis shows that the framework remains affordable, making it practical for real-world deployment.

\end{abstract}


\section{Introduction}

With the boom of artificial intelligence, a wide range of forecasting algorithms have been proposed recently, many of which have demonstrated impressive performance. However, these forecasting methods become static once designed, offering no mechanism for interaction between the model and human users. This lack of interaction creates major barriers to the practical use of the forecasting methods. The implementation of these algorithms involves complicated pipelines, including data cleaning, feature engineering, model selection, hyperparameter tuning, deployment, etc\cite{wang2023}. Successfully executing these processes requires substantial knowledge of programming, statistics, and machine learning. However, system operators who primarily focus on system operations rather than model development typically do not possess this knowledge. In addition, each forecasting task exists within a specific operational context. Operators are often aware of these context-specific factors and can provide intuitive judgments, domain insights, or scenario-dependent information that are difficult to capture from data alone. It is important to integrate the knowledge with the forecasting system to seamlessly match real-world decision-making needs, but effective approaches to achieve such a goal are still lacking.

These challenges highlight the need for an interactive load forecasting paradigm that can lower technical barriers to implementation and enable users to integrate their domain insights and contextual knowledge \cite{gil2019,mosqueira2023}. Building a bridge between users and the forecasting system requires an interface capable of understanding natural language, interpreting flexible instructions, and coordinating actions across the forecasting pipeline. Recent breakthroughs in LLMs have opened up new possibilities for intelligent, user-friendly interfaces across a wide range of technical domains. Trained on vast corpora of text and code, LLMs exhibit strong capabilities in processing natural language, reasoning over complex instructions, and orchestrating multi-step workflows. 

Beyond their general capabilities, LLMs have been adapted for domain-specific tasks through a variety of strategies, such as fine-tuning\cite{taylor2022}, reasoning enhancement\cite{wei2022,yao2023} and tool augmentation\cite{schick2023}. An emerging line of work introduces the notion of LLM-based agents—autonomous entities that perceive instructions, make decisions, and take actions\cite{wang2024(a)}. To support more complex or multi-stage tasks, recent studies have further explored multi-agent collaboration, where multiple agents interact and coordinate to solve problems beyond the capabilities of a single agent\cite{hong2024}.

Although the integration of LLMs into the power systems is still at an early stage, several emerging applications demonstrate their potential. Recent studies have explored using LLMs for correlation analysis, on-site hazard recognition, and document summarization tasks\cite{majumder2024}. Others have investigated possibilities in energy management\cite{zhang2025(b),michelon2025,yang2024}, power system analysis\cite{bonadia2023,jia2024}, anomaly detection\cite{zhang2025(b)} and optimal power flow problems\cite{yan2023,huang2024}. Despite these advancements, the application of LLMs in load forecasting remains relatively limited. Many existing studies focus on adapting LLMs to directly perform forecasting tasks, primarily through techniques such as prompt-based learning, time series patching, and fine-tuning the models on structured historical data\cite{liao2025,wang2024(b),zhou2025}. In contrast, much less attention has been given to leveraging LLMs as intelligent assistants that support users throughout the forecasting process. \cite{deng2024} proposed BuildProg, a tool that employs LLMs for generating Python programs to support testing of building load forecasting models via prompt engineering. \cite{zhang2025(c)} developed a LLM-based method that automates code generation for building energy load prediction. \cite{liu2025} developed RePower, an autonomous LLM-driven research platform that integrates data acquisition, algorithm design, and evolutionary optimization to automate data-driven power system tasks. These studies offer valuable insight into how LLMs can be embedded into research workflows, but they rely on rigid task structures, provide limited flexibility, and primarily position LLMs as static code generators or automation tools. Shifting from using LLMs as passive tools to enabling them to actively manage forecasting workflows and interact with users introduces significant challenges, including coordinating multi-stage processes without rigid programmatic control and handling user input in a context-aware manner.

Inspired by the aforementioned studies, we propose an LLM-based multi-agent collaborative framework to support interactive, real-world load forecasting tasks. We make the following contributions:
\begin{enumerate}
    \item We propose an LLM-based framework that enables human users to interact with the entire forecasting pipeline including preparation, model training, evaluation, and deployment. This framework reduce the technical barrier for non-expert users and allows users to express task-specific goals and expert knowledge using natural language.
    \item We design a multi-agent collaboration mechanism tailored to load forecasting, in which specialized agents are responsible for different stages of the forecasting pipeline and seamlessly communicate via a dedicated messaging system. This design enables effective execution by limiting task complexity of each agent while supporting flexible interaction with the user.
    \item We conduct extensive load forecasting experiments on two real-world datasets to evaluate the proposed framework. The results show that our framework effectively integrates user interaction into the forecasting pipeline, leading to improved predictive accuracy through informed human guidance. We also assess the efficiency of the proposed framework by analyzing token consumption. These findings demonstrate the practical feasibility of our framework in supporting interactive, high-quality forecasting in real-world power system scenarios.
     
\end{enumerate}


\section{Problem Statement}
\label{problem statement}

Load forecasting includes three main stages, namely task preparation, model training and evaluation, and model deployment. 

In the first stage, there are two core steps to be completed: data cleaning and task definition. Denote $\mathcal{D} = \{P_t, \Phi_t\}_{t=1}^T$ the raw data collected at the beginning of a forecasting task, where $P_t \in \mathbb{R}$ represents the historical load at time $t$ and $\Phi_t \in \mathbb{R}^d$ represents a $d$-dimensional vector of auxiliary features, such as calendar and weather information. Real-world data $\mathcal{D}$ often suffer from various imperfections, where a data cleaning process is needed. This process transforms the original data into a cleaned version $\tilde{\mathcal{D}}$ that is suitable for subsequent stages. Subsequently, the forecasting task is formally defined by specifying a forecasting interval $\Delta$, a forecasting horizon $H$, and an evaluation metric $\mathcal{L}$. Given historical observations up to time $t$, the goal is to predict the sequence of future load values $\boldsymbol{P}_t=\{P_{t+\Delta+1}, \dots, P_{t+\Delta+H} \}$. The generated prediction $\hat{\boldsymbol{P}}_t$ will be evaluated with the metric $\mathcal{L}$ which measures the discrepancy between the predicted and true load values. 

In the second stage, the goal is to explore different combinations of model types, hyperparameters and feature engineering options to identify the best configurations for forecasting. The model type $m$ is first selected from a set of commonly adopted candidates $\mathcal{M}_\text{types}$, where each model type $m$ is associated with model-type-specific hyperparameter space $\mathcal{H}^{(m)}$ and feature engineering options $f \in \mathcal{F}^{(m)}$. The overall search space $\mathcal{S}$ for the forecasting task is then defined as
\begin{equation}
    \mathcal{S} = \bigcup_{m \in \mathcal{M}_{\text{types}}} \{m\} \times \mathcal{F}^{(m)} \times \mathcal{H}^{(m)}, 
\end{equation}
where $\bigcup$ denotes the disjoint union over all model types, and $\times$ represents the Cartesian product between model selection, feature options, and hyperparameter configurations. The optimization objective is to search over $\mathcal{S}$ for the best configuration in terms of $\mathcal{L}$:
\begin{equation}
(m^{*}, f^{*}, h^{*}) = \underset{m \in \mathcal{M}_{\text{types}},\ f \in \mathcal{F}^{(m)},\ h \in \mathcal{H}^{(m)}}{\arg\min} \mathcal{L}(\hat{\boldsymbol{P}}_t, \boldsymbol{P}_t),
\end{equation}
where $(m^{*}, f^{*}, h^{*}) \in \mathcal{S}$ represent the optimal configurations. 

In the third stage, the best configuration obtained from the previous search is deployed for real-time forecasting. Let $\mathcal{D}_{\tau}$ denote the data used in online forecasting, which is first processed through the optimal feature engineering $f^*$ to generate the model input, and then passed to the trained model $m^*$ with hyperparameters $h^*$  to produce the forecast $\hat{\boldsymbol{P}}_{\tau}$.
Next, an optional postprocessing step can be applied to refine the prediction according to domain-specific requirements or expert experience. This postprocessing operation is represented as a transformation function $g(\cdot)$, which adjusts the model output to produce the final forecast $\tilde{\boldsymbol{P}}_{\tau} = g(\hat{\boldsymbol{P}}_{\tau})$.

The practical execution of above stages typically involves substantial manual design, which pose a high barrier for system operators. While LLMs offer promising capabilities for assisting such workflows, their use in supporting interaction in load forecasting processes remains largely unexplored. To this end, we identify the following two key research questions: 
\begin{itemize}
\item \textbf{Accessibility:} How can large language models serve as intelligent assistants to support users throughout the complex forecasting pipelines, enabling them to benefit from advanced models without requiring deep technical expertise?
\item \textbf{Human-in-the-Loop:} In what ways can human knowledge and preference be effectively incorporated into the forecasting process via LLM-based interaction?
\end{itemize}
With the aim of answering these questions, we provide detailed methodologies in the next section.

\section{Methodology}
\label{methodology}
In this section, we will introduce the proposed LLM-based interactive framework that is designed to not only help users handle complicated technical tasks, but also support users to actively participate throughout the entire process. We will first introduce the overall framework, covering the system architecture, agent roles, and communication mechanism. Then, we detail the design of each agent and explain the key technical components.

\subsection{Overall framework}
\begin{figure*}[htbp]
\centering
\includegraphics[width=0.8\linewidth]{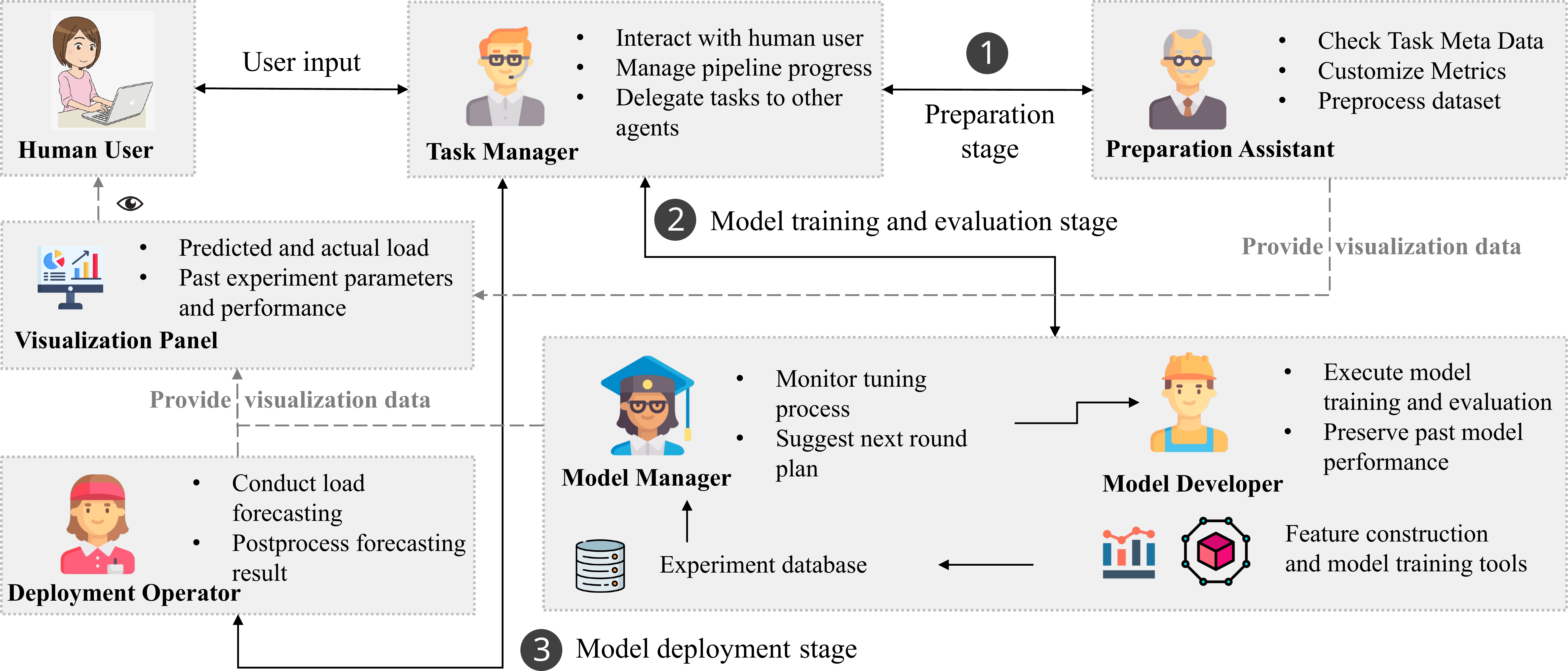}
\caption{The proposed multi-agent collaboration framework includes a Task Manager that handles user interaction and workflow control and other specialized agents manage different tasks in the pipeline.}
\label{proposed framework}
\end{figure*}
The proposed framework designs a multi-agent architecture, which has emerged as a promising approach for handling complex, multi-stage tasks\cite{hong2024,trirat2024}. As depicted in Fig.\ref{proposed framework}, the system consists of five specialized agents that collaboratively manage different stages of the forecasting pipeline. The key components are as follows:
\begin{itemize}
    \item \textbf{Task Manager:} Acts as the central coordinator of the system. It interacts with the human user, monitors the forecasting pipeline, and communicate with appropriate agents in each stage. 
    \item \textbf{Preparation Assistant:} Handles the initial preparation stage. This includes collecting task metadata, customizing evaluation metrics, and performing data preprocessing.
    \item \textbf{Model Manager:} Monitors the model optimization process. It keeps track of past performance information and incorporates user guidance to propose future exploration strategies.
    \item \textbf{Model Developer:} Executes training and evaluation of forecasting models based on configurations received from the Model Manager. 
    \item \textbf{Deployment Operator:} Deploys the best-performing model identified during the optimization stage, applies it to real-time data and performs postprocessing based on user preference.
\end{itemize}

In addition to the core agents, the framework includes a visualization panel that serves as an interface to present intermediate results that facilitate understanding and decision-making. These stage-aware visualizations are automatically triggered by internal events and continuously updated, where detailed visualized content will be introduced in the corresponding sections.

\begin{figure}[t]
\centering
\includegraphics[width=0.8\columnwidth]{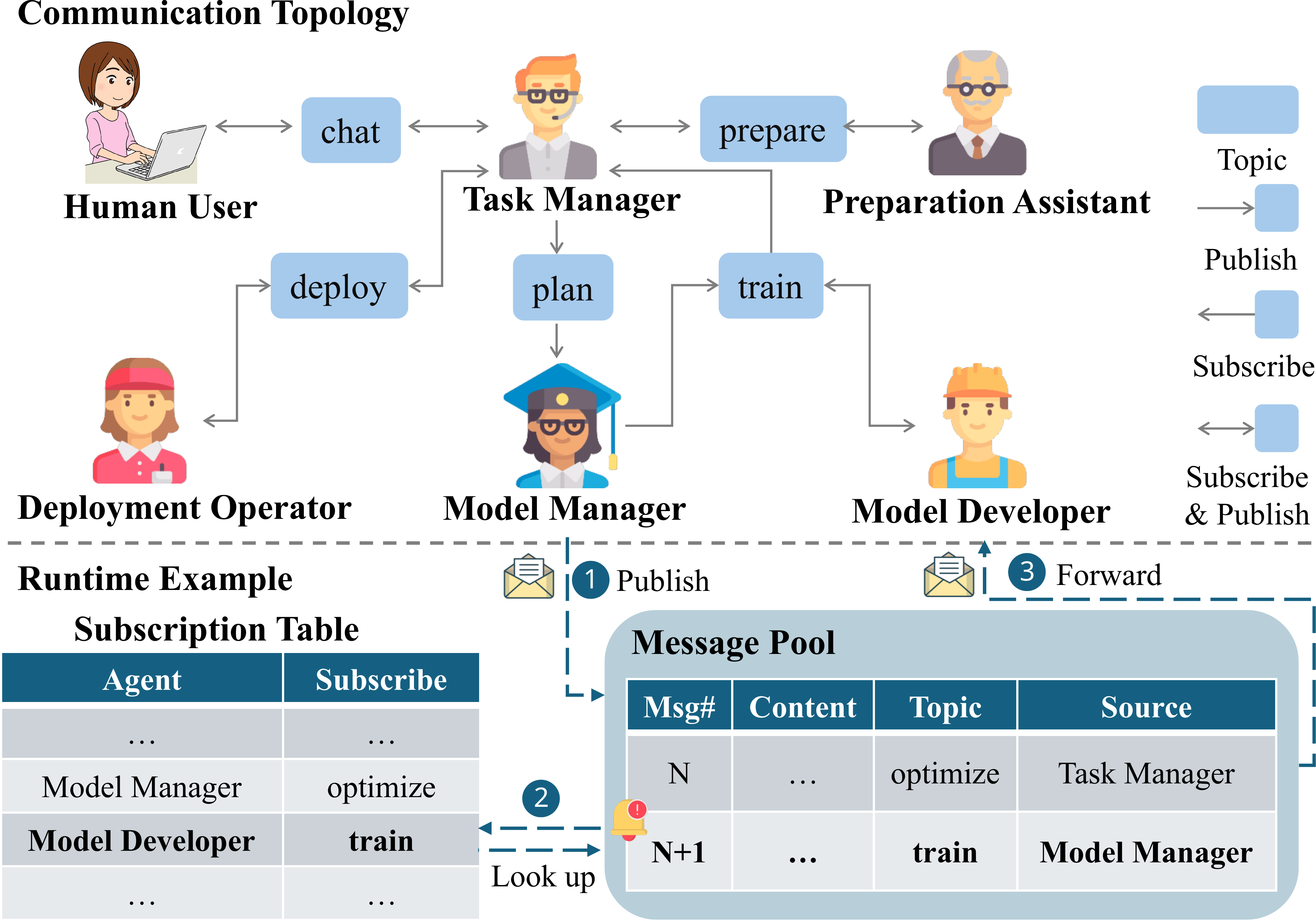}
\caption{Communication mechanism of the multi-agent framework, illustrating the topic-based messaging topology and a simplified example of the Model Manager sending a message to the Model Developer.}
\label{message routing}
\end{figure}

Moreover, we design a subscription-based message routing mechanism to coordinate communication between agents. As presented in Fig. \ref{message routing}, each agent registers at runtime with a predefined set of topics that reflect its responsibilities. Messages are published to a message pool along with their associated topics. During execution, agents monitor the pool and retrieve the messages corresponding to the subscribed topics. This design allows each agent to focus on its own scope of interest without requiring knowledge of other agents' internal logic.  

\subsection{Task Manager}
The Task Manager serves as the central coordinator of the multi-agent system. It is the only agent that interacts directly with the human user, acting as an abstraction layer that shields users from the underlying complexity of the execution pipeline.

\begin{figure*}[t]
\centering
\includegraphics[width=0.8\textwidth]{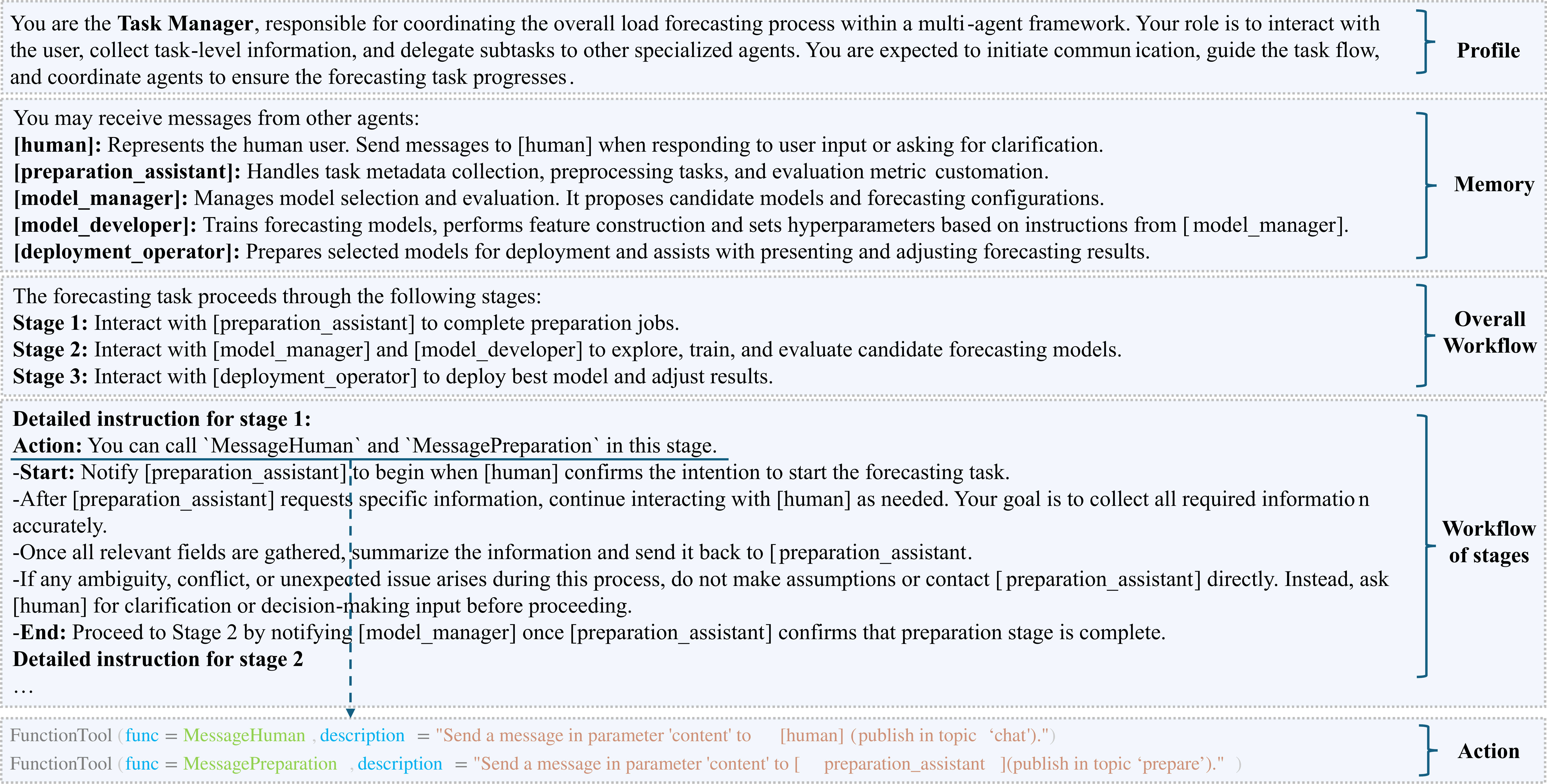}
\caption{A simplified system prompt of the Task Manager with four core components: profile, memory, workflow, and actions.}
\label{prompt example}
\end{figure*}

We use the Task Manager as a representative example to illustrate the design of a single agent in our framework. Each agent is defined through four key components: profile, memory, planning, and action \cite{luo2025}. We design a workflow-based approach instead of free-planning, where the expected behavior at each stage is explicitly described in the system prompt. As shown in Fig. \ref{prompt example}, the system prompt of the Task Manager clearly reflects these four components and provides a structured template for guiding agent behavior in context:

\begin{itemize}
    \item The \textbf{profile} section of the prompt defines the role and responsibilities of an agent. For the Task Manager, it should coordinate the entire forecasting pipeline, manage stage transitions and interact with the user.
    \item The \textbf{memory} section of the prompt specifies that each message is tagged with a role marker. This helps the agent distinguish between different message sources. In our system, the memory of each agent is constructed from the message history, where all incoming messages—regardless of source—are appended to a shared message buffer. 
    
    \item The \textbf{workflow} section defines the stage-wise behavior of the agent throughout the task. For each stage, the system prompt specifies what Task Manager should do, when to transition, and whom to communicate with. For conciseness, Fig. \ref{prompt example} only show detailed workflow for preparation stage, and this pattern continues across all stages.
    \item The \textbf{action} section of the system prompt defines the set of operations the agent is allowed to perform. For the Task Manager, this primarily involves sending messages to other agents or to the human user. These actions are implemented as tool calls, a mechanism that has become a widely supported and standardized component in modern LLM frameworks \cite{schick2023}. Other agents in the system are also equipped with a variety of functional tools tailored to their specific responsibilities.
\end{itemize}

The system prompts of the remaining agents follow the same structural design based on profile, memory, workflow, and action. In the following sections, we will not repeat these basic definitions, but instead focus on the unique aspects of each agent's behavior and responsibilities.

\subsection{Preparation Assistant}
\label{preparation assistant}
\begin{figure}[t]
\centering
\includegraphics[width=0.6\columnwidth]{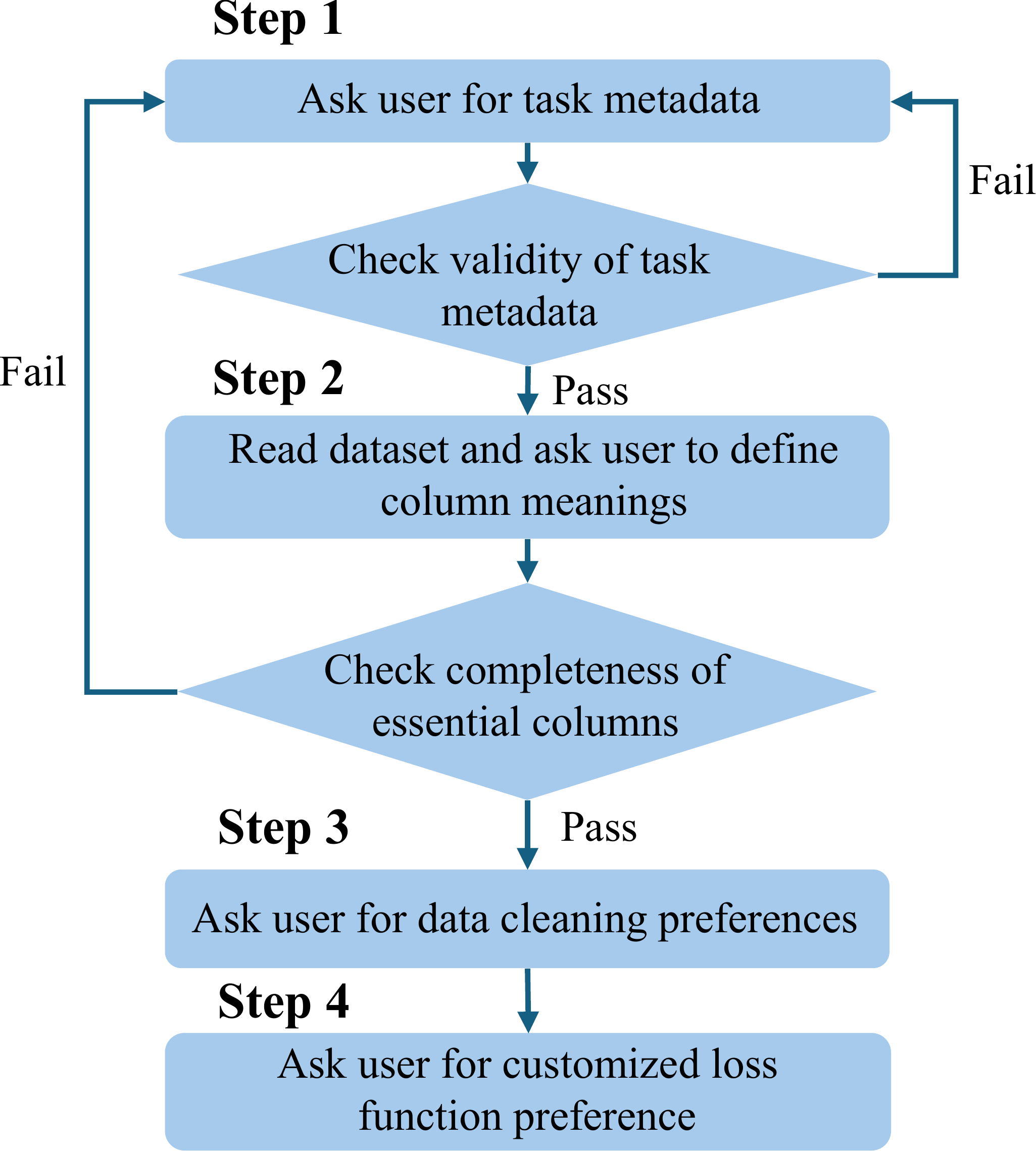}
\caption{Workflow of the task preparation stage. User interactions and system checks are conducted at each step to ensure correct task definition and data readiness.}
\label{preprocess steps}
\end{figure}

The Preparation Assistant is responsible for forecasting task preparation, which consists of four sequential steps as shown in Fig.~\ref{preprocess steps}. 

In \textbf{step 1}, the system prompts the user to provide essential task metadata, including the forecast interval $\Delta$, the forecast horizon $H$, and the path to the dataset. The system then checks the validity and completeness of the provided metadata before proceeding. During this process, the Preparation Assistant issues a request to the Task Manager, which in turn interacts with the user—looping as needed—until all required information is collected and returned. The same communication pattern is followed in all subsequent steps.

In \textbf{step 2}, the dataset is loaded and the semantic meaning of each column is identified based on user confirmation. The Preparation Assistant attempts to infer the meaning of each column and prompts the user to confirm or revise these assignments as needed. At a minimum, the dataset is required to contain a timestamp column and a load column. If these required fields are missing or incorrectly labeled, the system will revert to step 1 and request the user to provide a new dataset path.

\textbf{Step 3} involves a set of data cleaning procedures to identify anomalies and impute missing values via dedicated function tools \cite{wang2023}. In this step, the statistics of the data are presented on the visualization panel to help the user understand the overall structure of the data and the effects of applied cleaning operations. 

In \textbf{step 4}, the user is offered the options to customize the evaluation metric for the forecasting task. In addition to standard error metrics, three types of loss function customization strategies are supported to meet real-world needs. (i) The time-weighted loss assigns different importance to different time steps. In practice, a system operator may particularly care about the forecasting accuracy during certain periods, such as early evening hours when residential consumption surges. (ii) The condition-weighted loss adjusts the loss dynamically based on external conditions. For example, under high temperatures, air conditioning demand rises sharply, making accurate load forecasting more critical for maintaining system reliability. The above two weighted loss can be defined as:
\begin{equation}
\mathcal{L}_{\text{weighted}} = \frac{1}{\sum_{i} w_i} \sum_{i=1}^{H} w_i \cdot \ell(\hat{P}_{t+i}, P_{t+i}),
\end{equation}
where $w_i$ is the weight assigned to the $i$-th point based on its importance assigned by the user. (iii) The asymmetric loss captures the reality that the cost of over-predicting and under-predicting load are often not equal\cite{wang2017}. By specifying different penalty coefficients $\alpha$ and $\beta$, the forecasting model can be tuned to align with the real-world cost. The asymmetric loss can be formulated as:
\begin{equation}
\begin{aligned}
\mathcal{L}_{\text{asym}} = \frac{1}{H} \sum_{i=1}^{H} 
\begin{cases}
\alpha \cdot \ell(\hat{P}_{t+i}, P_{t+i}) & \text{if} \quad \hat{P}_{t+i} > P_{t+i}, \\
\beta \cdot \ell(\hat{P}_{t+i}, P_{t+i}) & \text{otherwise}.
\end{cases}    
\end{aligned}
\end{equation}

\subsection{Model Manager and Developer}
\label{model manager and developer}
\subsubsection{Algorithm Description}
In the model training and evaluation stage, the system aims to identify optimal combinations of model types, feature options, and model hyperparameters $(m,f,h)$, as described in Section~\ref{problem statement}. The problem falls within the scope of automated machine learning (AutoML), an actively researched area that has produced a wide range of algorithms, including Bayesian optimization, evolutionary strategies, and reinforcement learning-based approaches \cite{karmaker2021}\cite{barbudo2024}. Among these, Bayesian optimization has demonstrated strong performance and flexibility across a variety of configuration search problems, making it a widely adopted choice.

The standard Bayesian optimization algorithm optimizes the unknown objective function $\omega: \mathcal{S} \rightarrow \mathbb{R}$ in the search space $\mathcal{S}$. At iteration $n$, 
 it builds a probabilistic surrogate model $\hat{\omega}$ to approximate $\omega$ based on the past trials, and selects the next configuration by maximizing an acquisition function $a(\cdot)$ derived from $\hat{\omega}$\cite{watanabe2023}:
\begin{equation}
\begin{aligned}
&(m_{n+1}, f_{n+1}, h_{n+1}) \\= &\underset{(m, f, h) \in \mathcal{S}}{\arg\max} a\left((m, f, h) \mid \{(m_i, f_i, h_i, l_i)\}_{i=1}^{n}\right),    
\end{aligned}
\end{equation}
where $l_i$ denotes the performance metric obtained by evaluating $(m_i,f_i,h_i)$.

While this algorithm can search efficiently in standardized settings, the search space in our problem is hierarchical and highly non-uniform, as different model types involve distinct sets of hyperparameters and vastly different subspace complexities. In such a search space, purely acquisition-driven selection can lead to premature concentration on initially favorable model types. Therefore, instead of embedding the process into a monolithic AutoML module, we design an interactive Bayesian optimization algorithm that integrates both automated optimization and user-guided adjustment.

\begin{algorithm}[tb]
\caption{Proposed Interactive Bayesian Optimization Algorithm}
\label{alg:llm_bo}
\KwIn{Configuration search space $\mathcal{S}$, max trials $Tr$, target performance $\epsilon$}
\KwOut{Best configuration $(m^*, f^*, h^*)$}

Sample $K$ configurations $\{(m_i, f_i, h_i)\}_{i=1}^K \sim \textbf{RandomSample}(\mathcal{S})$\;
Evaluate $l_i = \omega(m_i, f_i, h_i)$ for $i = 1, \dots, K$\;
$\mathcal{R} \leftarrow \{(m_i, f_i, h_i, l_i)\}_{i=1}^K$\;

\Repeat{$|\mathcal{R}| \geq Tr$ or $\min l_i \leq \epsilon$}{
    $(\mathcal{S}', \mathcal{G}) \leftarrow \textbf{ExternalGuidance}(\mathcal{S}, \mathcal{R})$\;

    $\{(m_j, f_j, h_j)\}_{j=1}^{B} \leftarrow \underset{{(m, f, h) \in \mathcal{S}'}}{\arg\max} a\left((m, f, h) \mid \mathcal{R}, \mathcal{G}\right)$\;

    \ForEach{$(m_j, f_j, h_j)$ in batch}{
        Evaluate $l_j \leftarrow \omega(m_j, f_j, h_j)$\;
        $\mathcal{R} \leftarrow \mathcal{R} \cup \{(m_j, f_j, h_j, l_j)\}$\;
    }

    $\hat{\omega} \leftarrow \textbf{FitSurrogate}(\mathcal{R})$\;
}
$(m^*, f^*, h^*) \leftarrow \arg\min_{(m, f, h, l) \in \mathcal{R}} l$\;
\Return $(m^*, f^*, h^*)$
\end{algorithm}

As shown in Algorithm~\ref{alg:llm_bo}, the interactive algorithm begins with an initial random sampling of $K$ configurations and their evaluations, forming a result set $\mathcal{R}$. In each iteration of the searching process, external guidance obtained from user instructions and LLM reasoning is used to optionally prune the search space to $\mathcal{S}'$ or to impose preferences on the configurations of the next batch, denoted as the guidance context $\mathcal{G}$. A batch of $B$ candidate configurations is then selected via the acquisition function $a(\cdot)$ conditioned on $\mathcal{R}$ and $\mathcal{G}$, evaluated using the objective $\omega(\cdot)$, and added to the result set. The surrogate model $\hat{\omega}$ is updated accordingly. This loop continues until the stopping condition is met, after which the best configuration is returned.

This design preserves the efficiency of standard Bayesian optimization algorithms and leverages user insights obtained from observing visualized intermediate results. 

\subsubsection{Agent Design}
\begin{figure*}[t]
\centering
\includegraphics[width=0.8\textwidth]{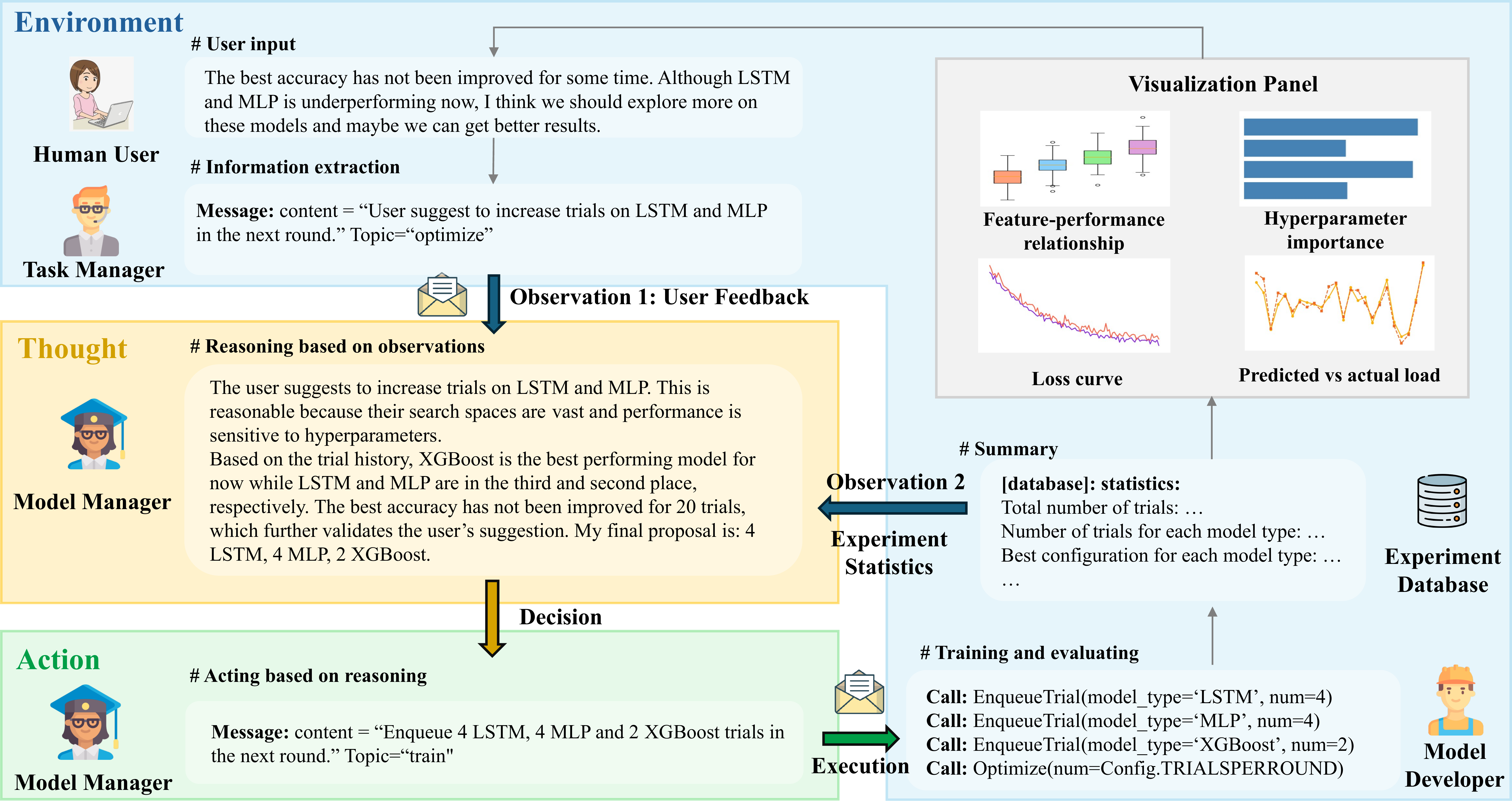}
\caption{Example workflow of the model training and evaluation stage. The user's input is first filtered and processed by the Task Manager, then forwarded to the Model Manager. The Model Manager follows a two-stage reasoning and action process before passing the resulting decision to the Model Developer, who is responsible for executing the corresponding function calls to train and evaluate the model. The evaluation results are then visualized and presented to the user.}
\label{fig:workflow of MM and MD}
\end{figure*}

The proposed algorithm is coordinated by two specialized agents: the Model Manager, which is responsible for searching strategy planning, and the Model Developer, which executes the corresponding training and evaluation tasks. A simplified runtime snapshot of one optimization iteration (line 5-10 in Algorithm~\ref{alg:llm_bo}) is presented in Fig.~\ref{fig:workflow of MM and MD}, which is drawn from the perspective of the Model Manager, highlighting how it interacts with the other agents and incorporates user input.

The Model Manager operates as a planning agent that follows a loop similar to the ReAct paradigm\cite{yao2023}. Apart from the instructions from user, it also receives a text summarization of the past trials, such as the number of evaluations per model type, best-performing configurations, and recent performance trends. Based on these observations, the Model Manager goes through a two-stage decision process. In the thinking stage, it analyzes the collected information to assess the relative potential performance of different model types, monitor the balance between exploration and exploitation, and interpret the guidance from user. When the user offers no guidance, the Model Manager falls back to a default strategy: based on accumulated performance statistics, it decides whether to allocate the next round of trials across model types to improve exploration balance, or simply allow the underlying optimization algorithm to explore freely without intervention. After that, in the action stage, the Model Manager translates its decision into a concrete set of configuration-level instructions, which are passed to the Model Developer via the message-based function call.

The Model Developer acts as a dedicated executor that receives trial plans from the Model Manager and transforms them into tool-level instructions.  For each trial, it invokes the appropriate feature engineering and model training functions and collects evaluation results. These functions are pre-implemented and encapsulated with a unified interface, which allows the system to remain modular and extensible. New models or feature engineering methods can be easily integrated without modifying the agent logic. This design also improves the robustness of execution, as the Model Developer can avoid dealing with method-specific implementation details. While the Model Developer currently acts mainly as a tool dispatcher, its separation from the Model Manager is essential. If the Model Manager were responsible for both planning and executing trial runs, it would need to embed numerous function calls within its prompts, which introduces significant complexity, increases the risk of generation errors, and potentially distracts the LLM from higher-level reasoning.

After a batch of selected configurations is evaluated, users are provided with visual feedback through the visualization panel. This includes global views such as plots of feature-performance relationships and estimates of hyperparameter importance which are derived from past evaluation results. For individual configurations, the system also presents training and validation loss curves, as well as true-versus-predicted load plots to help users spot performance bottlenecks and diagnose potential issues like underfitting or overfitting.

Based on this feedback, users can provide suggestions or preferences to guide the next round of exploration. The system allows users to influence the search through two primary mechanisms. (i) User can specify constraints or priorities for the next batch of trials. Apart from adjusting the allocation of trials across different model types similar to the Task Manager, users are also enabled to directly insert specific configurations (either partially or fully specified) for evaluation. This design ensures that users retain absolute control over the search direction when needed. (ii) User can also prune the search space for all subsequent trials, such as narrowing parameter ranges or excluding model types that appear unpromising. User's intentions are conveyed in natural language inputs, and are interpreted by the Task Manager and mapped into the aforementioned two mechanisms. The extracted instructions are then forwarded to the Model Manager again and incorporated into subsequent iterations of the optimization process.

\subsection{Deployment Operator}
\label{deployment operator}
After completing configuration search and model training in Stage 2, the best configuration $(m^{*}, f^{*}, h^{*})$ is deployed to forecast in a real-time environment. Provided with the new data, the Deployment Operator invokes the trained model to generate forecasts and presents forecasting results on the visualization panel, together with relevant input features such as temperature or calendar information.

When the initial forecasts are produced, the framework allows users to adjust the results by applying post-processing operations. The Deployment Operator can support a range of postprocessing strategies, among which we consider four commonly used types in this work, as shown in Table~\ref{tab:postprocess_option}. (i) Users may directly override specific forecast values through manual specification. (ii) Forecasts can be scaled within selected time intervals, providing intuitive corrections for scenarios involving special events or anticipated anomalies. In addition, scaling can be conditionally triggered when (iii) predicted load values or (iv) external variables (e.g., temperature) exceed certain thresholds. These two postprocessing strategies enable users to refine forecasts in cases where the model may systematically over- or under-predict. The framework is designed to be extensible, allowing additional strategies to be integrated as needed.

The Deployment Operator can be readily extended to support advanced functionality. The visualization module, for example, can be enhanced by incorporating historical user adjustments. Over time, the system could record when and how users modified forecasts, such as the extent of upward corrections during heat waves, the associated prediction errors, and the context in which adjustments were made. These historical patterns could then be displayed on the visualization panel to assist users in making more informed decisions in similar future scenarios. The Deployment Operator can also serve as an assistant for model management by interfacing with a library of forecasting models trained during past usage. With appropriate designs, it could help users assess whether a selected model remains suitable for current data or whether updates should be triggered.

\begin{table*}[!t]
\centering
\caption{Postprocessing options in model deployment stage}
\begin{tabularx}{\textwidth}{>{\raggedright\arraybackslash}p{0.2\textwidth}>{\raggedright\arraybackslash}p{0.25\textwidth}>{\raggedright\arraybackslash}p{0.2\textwidth}>{\raggedright\arraybackslash}p{0.3\textwidth}}
\toprule
\textbf{Type} & \textbf{Description} & \textbf{Expression} & \textbf{Application Scenario} \\
\midrule
\textbf{Manual Override} & Set predicted load at a specific hour to a user-specified value & $\hat{P}_t \leftarrow v_t, \quad t \in \mathcal{T}$ & When the user has strong prior beliefs or operational intentions regarding the expected load \\
\textbf{Time-Based Scaling} & Multiply predicted values in a time interval by a factor & $\hat{P}_t \leftarrow (1+\lambda)\hat{P}_t, \quad t \in \mathcal{T}$ & When a time segment is expected to deviate from normal patterns \\
\textbf{Load-Based Scaling} & If the predicted load exceeds a threshold, apply adjustment & $\hat{P}_t \leftarrow (1+\lambda)\hat{P}_t \quad \text{if} \quad \hat{P}_t \gtrless \theta$ & When under-/over-prediction risk increases at high or low load levels \\
\textbf{External-Variable-Based Scaling} & If an external indicator exceeds a threshold, apply adjustment & $\hat{P}_t \leftarrow (1+\lambda)\hat{P}_t \quad \text{if} \quad \phi_t \gtrless \theta$ & When external conditions (e.g. temperature, event signal) cause increased under-/over-prediction risk \\
\bottomrule
\end{tabularx}

\label{tab:postprocess_option}
\end{table*}

\section{Case Studies}
\label{case studies}
In this section, we present experimental results to analyze and validate the proposed framework. We begin by detailing the experimental setup, including the implementation specifics, the datasets used, and the model configurations. Next, we evaluate the role of human interaction in different aspects. Finally, we provide an analysis of the cost in terms of token consumption.

\subsection{Experimental Setups}
\subsubsection{Implementation Details}
In the case studies, we investigate the 24-hour-ahead electricity load forecasting, with $H=1$ and $\Delta=24$. We have tested several large language models (LLMs), including models from OpenAI, Anthropic and DeepSeek. Our framework is designed to be compatible with different LLM backends. In this case study, we use GPT-4o\cite{hurst2024}, accessed via the OpenAI API, as a representative example due to its strong performance in reasoning, instruction following, and tool use.
For the communication mechanism between agents, we leverage AutoGen\cite{wu2024}, an open-source multi-agent framework that provides convenient abstractions for message routing, agent registration, and execution management. For the interactive Bayesian Optimization algorithm in model training and evaluation stage, we use the python package provided by Optuna\cite{akiba2019}, which offers a highly integrated and flexible interface for defining complex search spaces and supports easy insertion of custom trials. We set the batch size to $B=10$ in the interactive model optimization stage, and enable the user to decide at the beginning of each iteration whether and how to intervene in the search process. 

\subsubsection{Data Description}
This study uses two real-world datasets to evaluate the proposed framework. The first dataset is from the Global Energy Forecasting Competition 2014 (GEFCom2014) load forecasting track\cite{hong2016(b)}. It contains hourly electricity load data and temperature data, where data from year 2013 and 2014 is used in our experiments. The second dataset consists of hourly electricity load and temperature data from Guangdong Province, China, spanning the years 2022 and 2023. For Guangdong dataset, in addition to time, load, and temperature, we include indicators of holidays as well as other weather information such as humidity and precipitation.

\subsubsection{Feature and Model Settings}
We consider a diverse set of forecasting models, including multiple linear regression (Linear), support vector regression (SVR), multilayer perceptron (MLP), XGBoost, and several deep learning architectures including LSTM, GRU, and CNN. For each model type, we define a dedicated hyperparameter search space $\mathcal{H}$ as summarized in Table~\ref{tab:hyperparameter_search}. 
Depending on the nature of each hyperparameter, we apply appropriate sampling strategies such as categorical choices, uniform sampling, log-uniform sampling, or discrete-step sampling. 

\begin{table}[]
\caption{Search space $\mathcal{H}$ for model hyperparameters}
\renewcommand{\arraystretch}{1.1}
\begin{tabular}{l|lll}
\hline
\textbf{Model Type} & \textbf{Hyperparameters} & \textbf{Search Space} & \textbf{Sampling Method} \\ \hline
\multirow{2}{*}{Linear} & regularization & \{none, ridge\} & categorical \\
 & alpha & {[}0.0001,1{]} & log-uniform \\ \hline
\multirow{2}{*}{SVR} & C & {[}0.001,1000{]} & log-uniform \\
 & gamma & {[}0.001,1000{]} & log-uniform \\ \hline
\multirow{4}{*}{MLP} & hidden layers & {[}2,5{]} & step=1 \\
 & hidden size & {[}16,512{]} & log-uniform \\
 & learning rate & {[}0.0001, 0.1{]} & log-uniform \\
 & dropout & {[}0,0.5{]} & uniform \\ \hline
\multirow{3}{*}{XGBoost} & n estimators & {[}10,300{]} & step=10 \\
 & max depth & {[}4,16{]} & step=2 \\
 & learning rate & {[}0.0001, 0.1{]} & log-uniform \\ \hline
\multirow{4}{*}{LSTM/GRU} & layers & {[}1,3{]} & step=1 \\
 & hidden size & {[}16,512{]} & log-uniform \\
 & FC layer size & {[}16,512{]} & log-uniform \\
 & learning rate & {[}0.0001, 0.1{]} & log-uniform \\ \hline
\multirow{5}{*}{CNN} & layers & {[}1,3{]} & step=1 \\
 & kernel size & {[}1,5{]} & step=1 \\
 & filters & {[}16,128{]} & log-uniform \\
 & FC layer size & {[}16,512{]} & log-uniform \\
 & learning rate & {[}0.0001, 0.1{]} & log-uniform \\ \hline
\end{tabular}
\label{tab:hyperparameter_search}
\end{table}
To control the complexity of the configuration space, we organize feature construction options into several groups, as summarized in Table~\ref{tab:feature_search_space}. Calendar-related features can be included in one of several encodings—numerical, categorical, or trigonometric—or excluded entirely. Following the setup in \cite{wang2016}, we construct the full candidate set of temperature-related features by including hourly temperature lags over the past 72 hours, as well as daily average temperatures for the past 3 days. We compute the Pearson correlation between each of them and the target load, rank these features accordingly, and retain the top proportion based on a tunable threshold. Interaction terms between calendar and temperature features are optionally included\cite{hong2011}. For Linear, SVR, MLP and XGBoost, the full candidate set of lagged load values consists of the past 168 hourly observations and is also selected via correlation. We additionally include a fixed selection strategy that incorporates the load values from the same hour over the past seven days \cite{qin2024}. For LSTM, GRU and CNN, the input load sequence is defined by its sampling frequency and length.
\begin{table}[!t]
\centering
\caption{Search space $\mathcal{F}$ for feature construction options}
\renewcommand{\arraystretch}{1.3}
\begin{tabularx}{\columnwidth}{>{\raggedright\arraybackslash}p{0.5\columnwidth}>{\raggedright\arraybackslash}X}
\hline
\textbf{Feature Group} & \textbf{Search Space} \\ \hline
Calendar Features & \{none, numerical, categorical, trigonometric\} \\ \hline
Temperature Lag Values & \{none, correlation\} \newline top-ratio if 'correlation' chosen: [0, 1] \\ \hline
Calendar-Temperature Interaction & \{none, all\} \\ \hline
Load Lag Values (For Regression Models) & \{none, correlation, fixed\} \newline top-ratio if 'correlation' chosen: [0, 1] \\ \hline
Historical Load Sequence (For Non-Regression Models) & frequency: \{1,2,3,4,6,12,24\} \newline length: [1,24] \\ \hline
Other Features & \{none, correlation\} \newline top-ratio if 'correlation' chosen: [0, 1] \\ \hline
\end{tabularx}
\label{tab:feature_search_space}
\end{table}

\subsubsection{Evaluation Metrics}
In our experiments, the optimization uses standard mean absolute error (MAE) as objective. Therefore, the results in the remaining part are presented using MAE as the primary evaluation metric. Additionally, we report mean absolute percentage error (MAPE) to offer a more intuitive view of forecasting performance across the two datasets.

\subsection{Effect of Human Interaction}
\subsubsection{Model Optimization}
This part presents a case study to illustrate how human interaction can influence the model optimization process. By analyzing the search trajectory with and without user intervention, we show that timely and proper guidance can improve both the effectiveness and efficiency of the search process.

Table \ref{tab:auto-vs-guided} presents a comparison between the vanilla Bayesian optimization baseline and the algorithm proposed in Section \ref{model manager and developer} across the two datasets. For the vanilla method, each value is the average of five independent trials, including both the best metrics achieved and the number of trials taken to reach them. The proposed method outperforms the baseline on both datasets and also reaches its best result using fewer trials, demonstrating not only improved accuracy but also higher efficiency. 

\begin{table}[]
\centering
\caption{Comparison of vanilla and proposed methods across datasets and error metrics}
\renewcommand{\arraystretch}{1.2}
\begin{tabular}{ccccc}
\hline
\multirow{2}{*}{Dataset} & \multirow{2}{*}{Metric} & \multicolumn{2}{c}{Method} & \multirow{2}{*}{Trials} \\ \cline{3-4}
 &  & Automated & Proposed &  \\ \hline
\multirow{2}{*}{GEFCom14} & MAE & 73.41 & 68.23 & 221 \\
 & MAPE & 2.24\% & 2.06\% & 183 \\ \hline
\multirow{2}{*}{Guangdong} & MAE & 2119.88 & 1966.96 & 218 \\
 & MAPE & 3.08\% & 2.96\% & 165 \\ \hline
\end{tabular}
\label{tab:auto-vs-guided}
\end{table}

\begin{figure}[!t]
    \centering
    \subfloat[\small Search process of vanilla Bayesian Optimization algorithm.]{%
        \includegraphics[width=0.83\columnwidth]{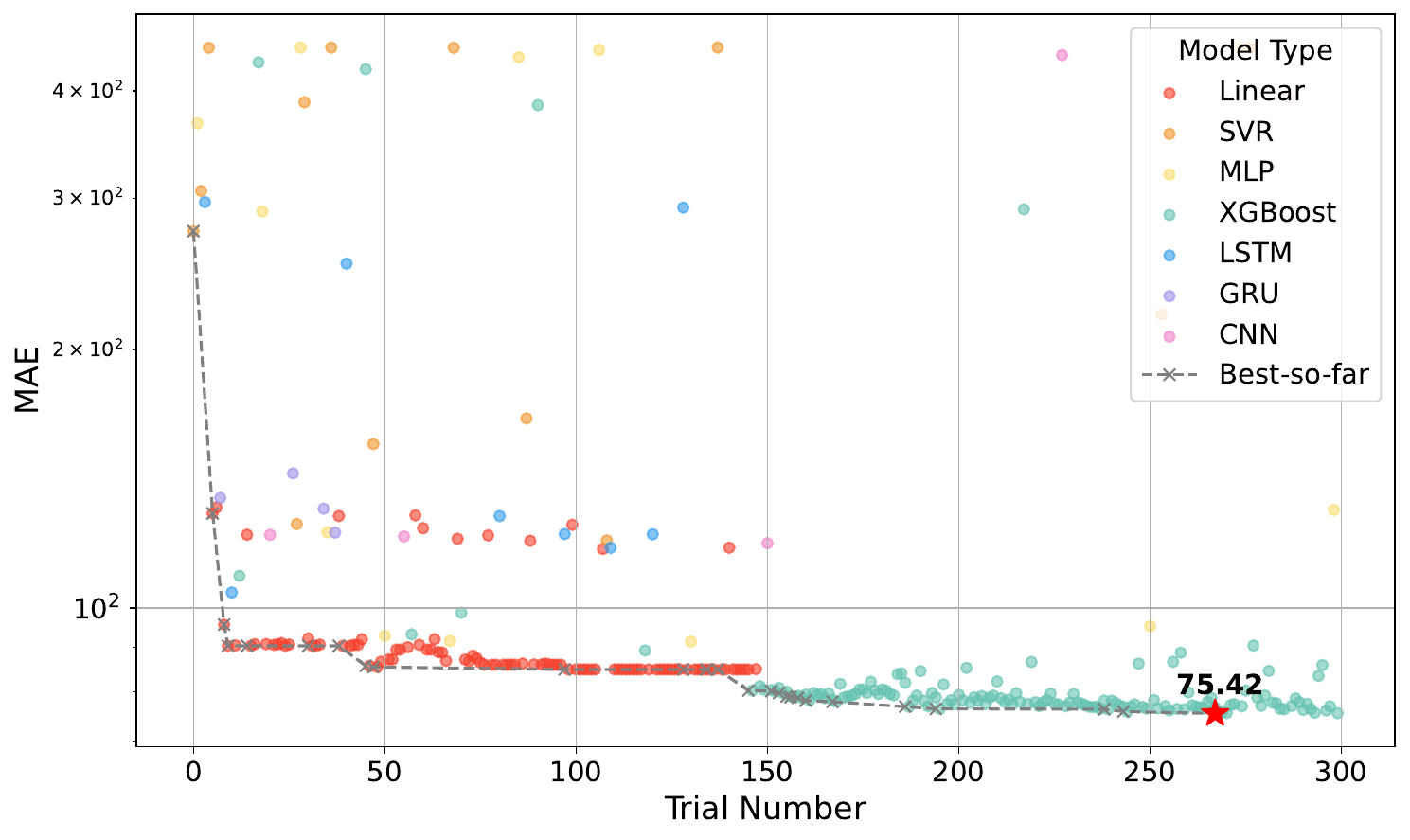}
        \label{fig:auto-search}
    }
    \hfill
    \subfloat[\small Search process of the proposed interactive algorithm.]{%
        \includegraphics[width=0.83\columnwidth]{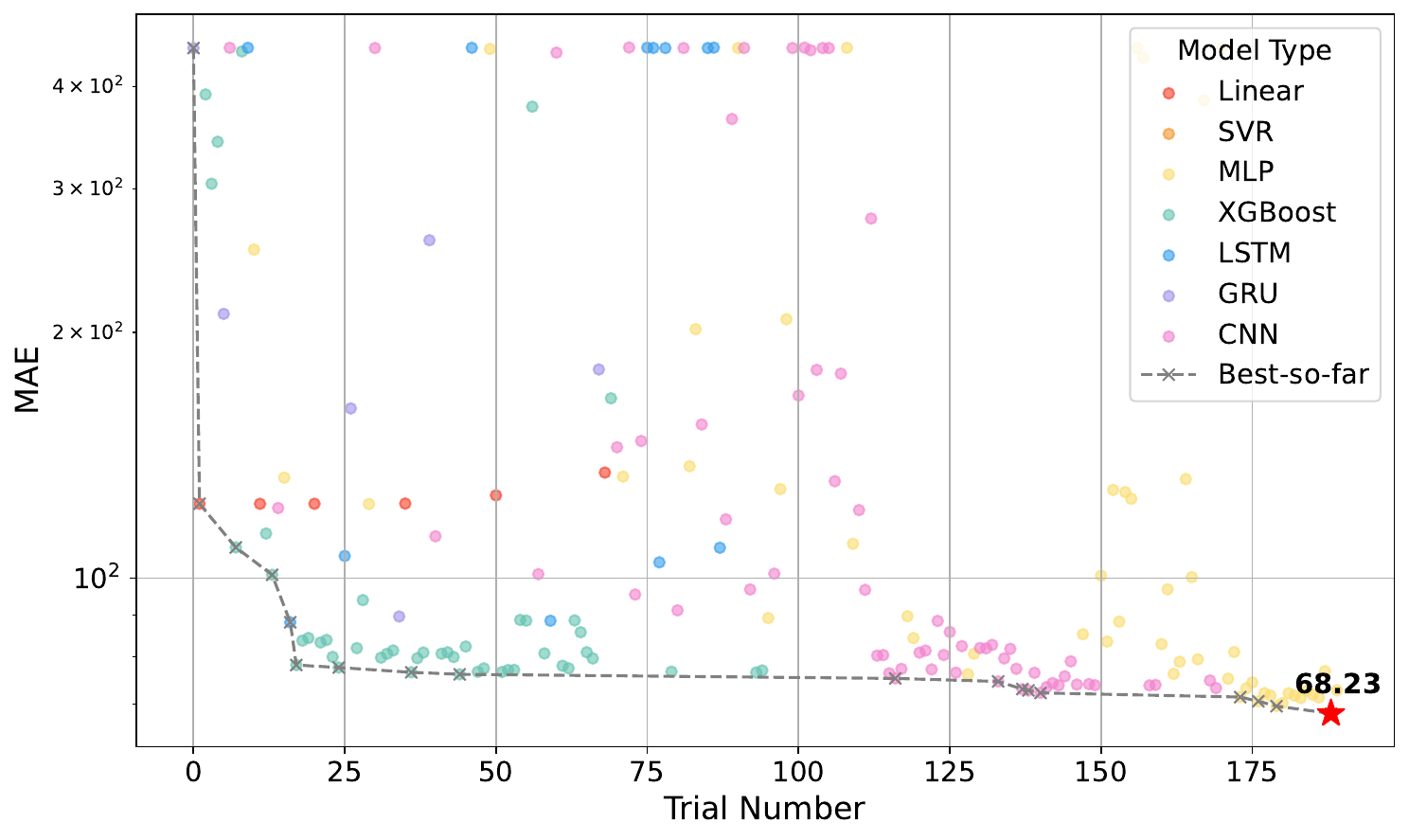}
        \label{fig:guide-search}
    }
    \caption{Examples of the detailed search trajectories for the baseline and proposed algorithms. Different model types are represented by color-coded scatter points, and the best-so-far curve along with the final optimal result is marked. For clarity of presentation, Figure~(b) is truncated after trial 190, as the search shows no improvement beyond this point. }
    \label{fig:seach-compare}
\end{figure}

To further illustrate the reason behind the performance gap, we present a comparison of detailed searching process of both methods on GEFCom14 dataset. Fig.~\ref{fig:auto-search} shows the scatter plot of one of the search processes conducted by the baseline method. During the early stage of the search, a relatively good result was found by chance using the Linear model. As a result, the optimization process focused heavily on this model type, and the performance plateaued for an extended period. It was not until around trial 150 that a better configuration involving XGBoost was discovered. Subsequently, the algorithm shifted its focus to XGBoost, but the performance improved only slowly until the maximum number of trials was reached.

Fig. \ref{fig:guide-search} illustrates the search process of the proposed method with human-in-the-loop interaction. In the early stages, the algorithm identified XGBoost as a promising model and focused primarily on exploring its configurations. However, a performance plateau was soon encountered. Around the sixth iteration (trial 60), the user observed stagnation in accuracy and revisited several previously underexplored but moderately performing models, including LSTM, CNN, and MLP. In subsequent rounds, these alternative model types were selectively explored. By iteration 10, based on intermediate results, the user chose to prioritize CNN while continuing limited exploration of MLP. Around iteration 15 (trial 150), it became evident that CNN had also reached a performance ceiling—although by this point the accuracy already surpassed that of the automated method. Moreover, recent CNN trials showed minimal variation in configurations. The user then shifted the main search direction to MLP, which led to a significant improvement in the final rounds.

Notably, the user also made adjustments during the search that accelerated convergence. For example, when exploring CNN, the learning rate was constrained to values below 0.001 based on prior results. Similarly, when exploring MLP, the user excluded 'none' and 'categorical' options from calendar feature encodings, focusing on more informative representations. 

The proposed interactive mechanism supports users with basic understanding of AI techniques, allowing them to guide the process and potentially achieve better results. For users with no experience in this area, the automated algorithm remains capable of delivering reasonably good performance without requiring manual intervention. The framework is thus flexible and inclusive: users at different levels of AI proficiency can engage with the system according to their capabilities and obtain results corresponding to their level of involvement.

\subsubsection{Postprocessing}
In this part, we present two examples to demonstrate the importance of human interaction in the model deployment stage. 

\begin{figure}[!t]
    \centering
    \subfloat[\small Predicted, Actual and Adjusted Load on Memorial Day, 2014 (GEFCom14).]{%
        \includegraphics[width=0.8\columnwidth]{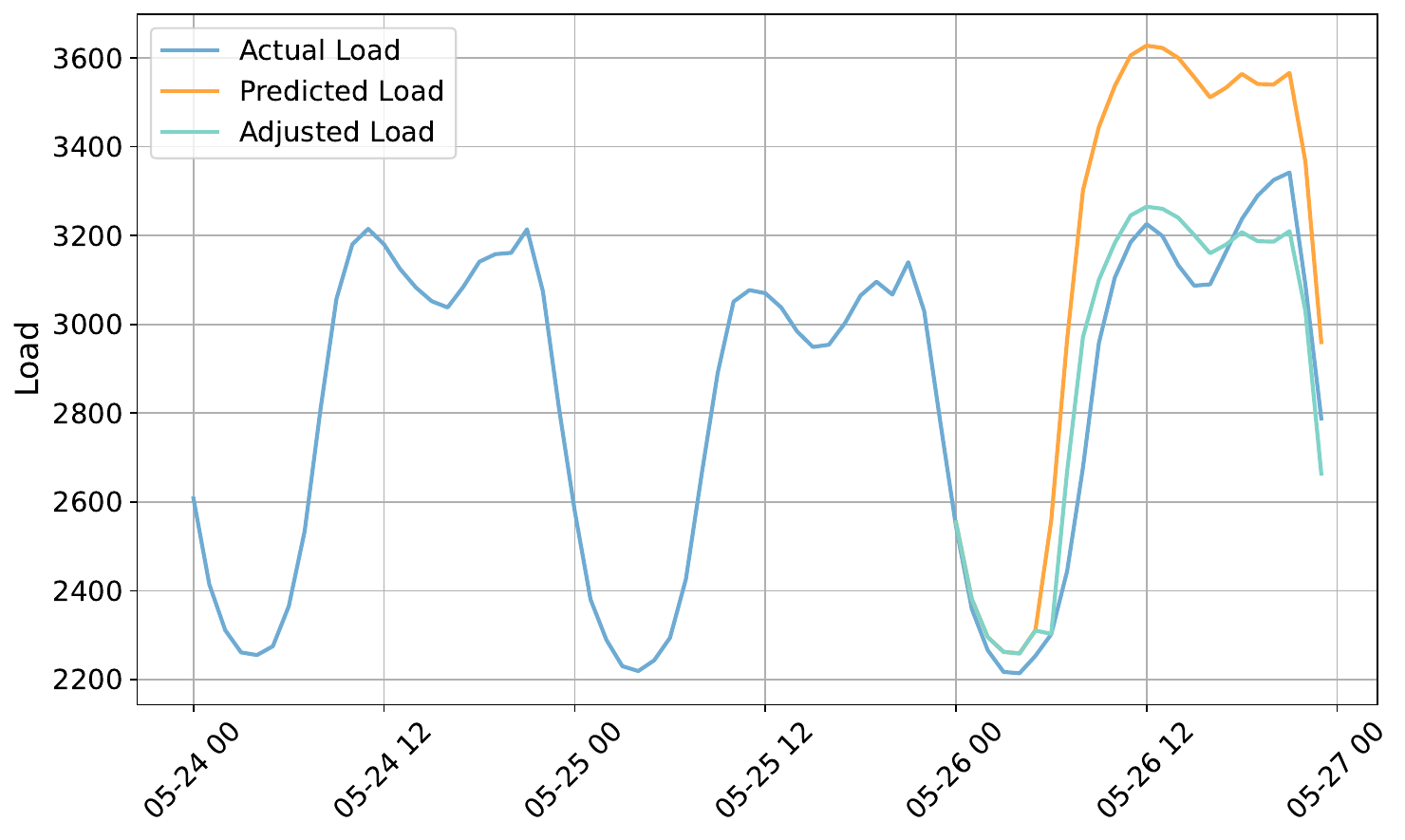}
        \label{fig:post-gef}
    }
    \hfill
    \subfloat[\small Predicted, Actual and Adjusted Load during Typhoon Saola, 2023 (Guangdong).]{%
        \includegraphics[width=0.8\columnwidth]{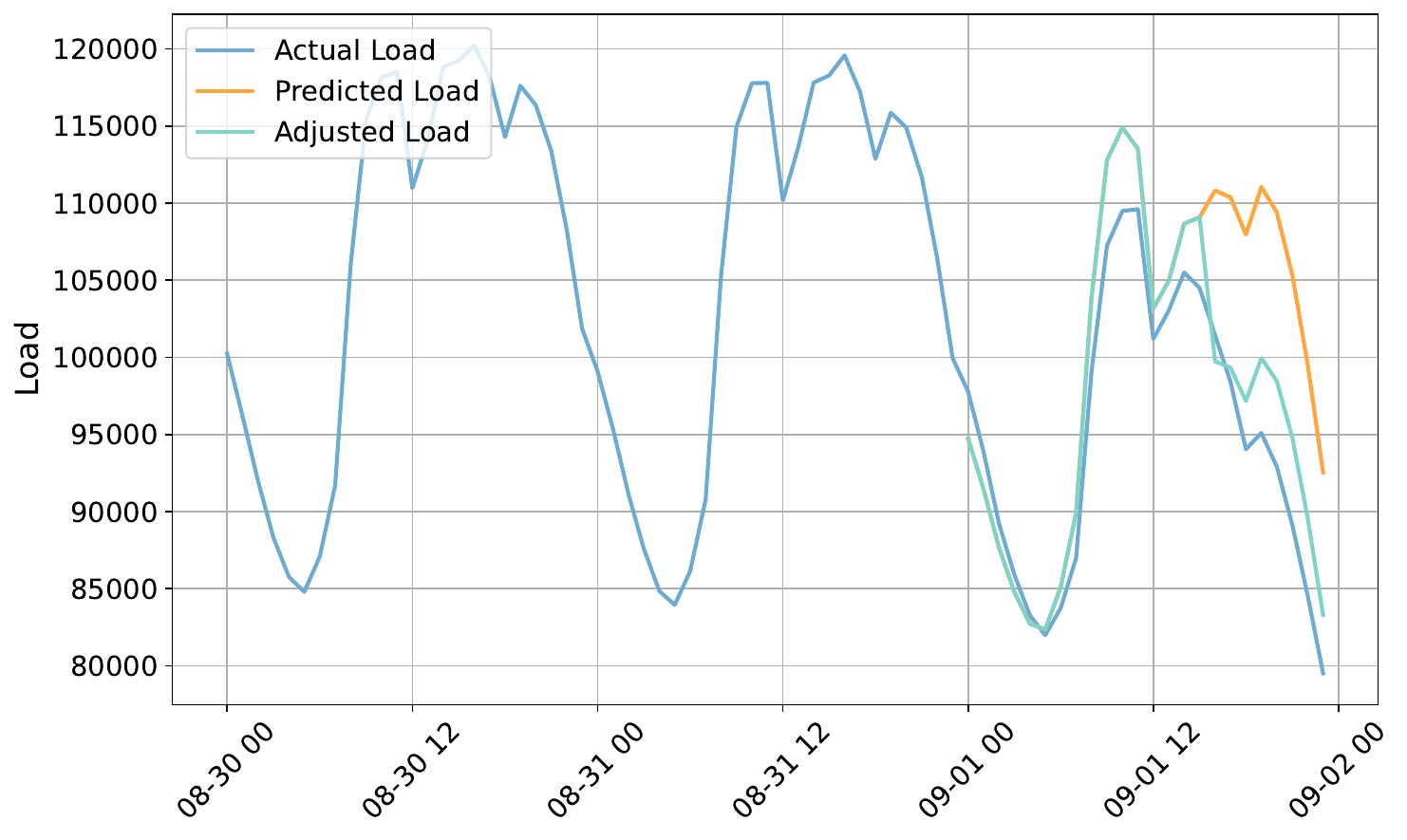}
        \label{fig:post-gd}
    }
    \caption{Users can postprocess the predicted load leveraging the situational context and their experience to enhance forecasting accuracy.}
    \label{fig:event_analysis}
\end{figure}

The first case uses data from the GEFCom2014 dataset. On 26 May 2014 (Memorial Day), the model overestimates the load during daytime hours because holiday information is not included in the raw dataset, and the predicted load follows the pattern of a regular Monday. The user is aware of typical demand reductions on holidays, and may manually reduced the forecast after 6 a.m. by 10\%, leading to improved alignment with actual load as shown in Fig.\ref{fig:post-gef}. 

The second case, based on the Guangdong dataset, presents a more challenging scenario. On September 1st, 2023, Typhoon Saola struck the region and significantly disrupted normal activity after 15:00. As illustrated by Fig.\ref{fig:post-gd}, although the model was trained with weather information, it failed to capture the actual impact of the typhoon on electricity demand. A user adjustment that reduced the forecast by 10\% after 15:00 offered a more reasonable reflection of the typhoon’s effect on social activity and electricity consumption, reducing MAPE from 6.99\% to 3.28\%. Such scenarios are common in real-world deployment but cannot be fully learned by data-driven models from historical data.

These two examples highlight the value of user intervention in scenarios where the model lacks sufficient contextual information. As introduced in Section \ref{deployment operator}, the framework can be further enhanced to better support users—particularly system operators—by integrating historical adjustment records and scenario retrieval mechanisms. When facing high-impact events, users could benefit from being shown similar past situations, how adjustments were made, and what outcomes followed.

\subsection{Cost Analysis}
\begin{figure}[t]
\centering
\includegraphics[width=0.85\columnwidth]{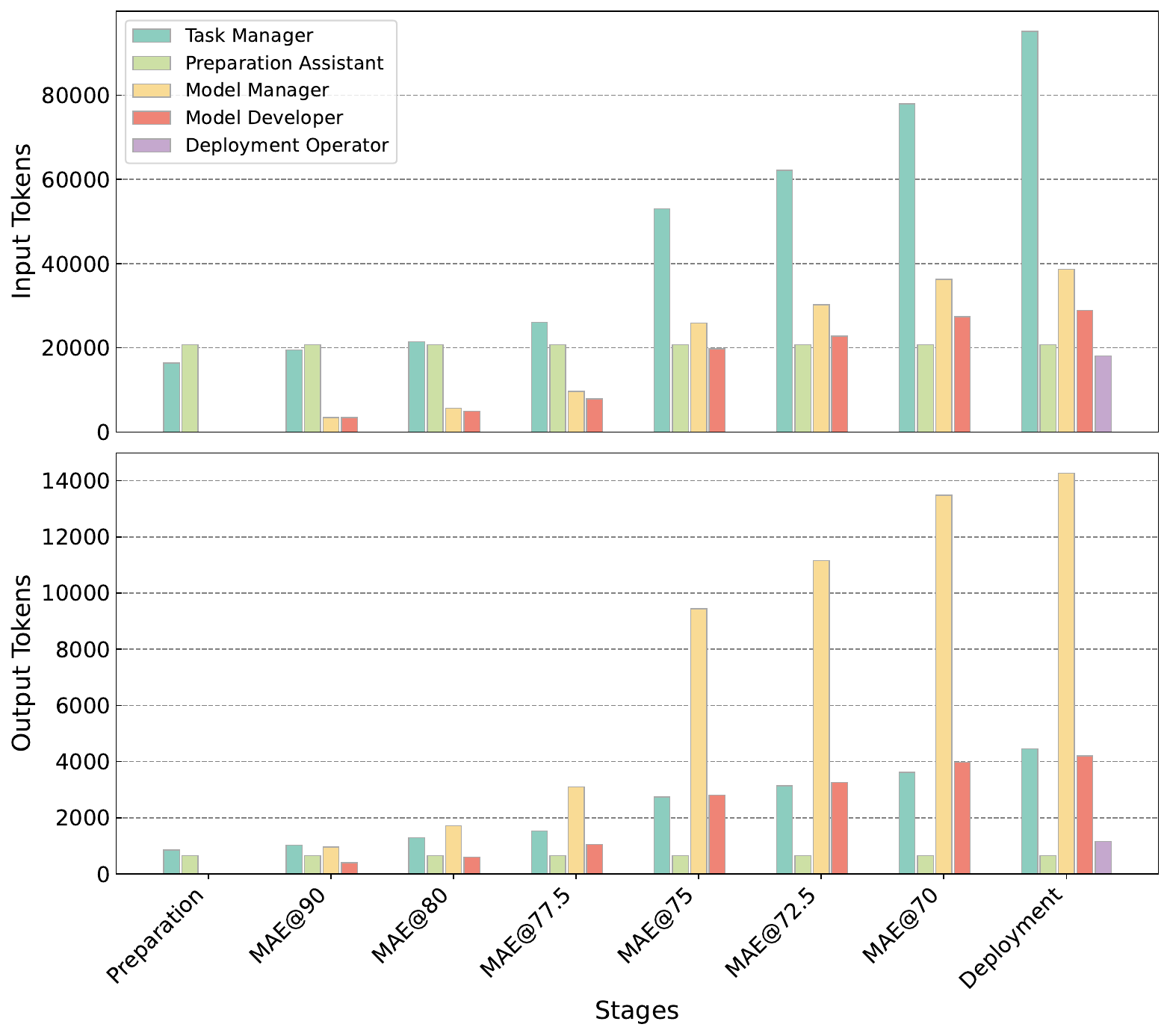}
\caption{Cumulative token cost analysis of agents in different stages of the pipeline.}
\label{fig:token-cost}
\end{figure}

To understand the cost of using the proposed framework, we analyze token usage across different stages of the forecasting workflow. Since token consumption is largely independent of the dataset, we report results based on the GEFCom14 dataset as examples. In Fig. \ref{fig:token-cost}, token consumption at each stage is reported cumulatively. For model training and evaluation stage, the bars marked with MAE values represents the total number of tokens consumed by the agents up to the point where the specified MAE level is first achieved.

During the preparation stage, the Task Manager and Preprocess Assistant exhibit similar levels of input and output token consumption, with input tokens significantly exceeding output tokens. This is primarily due to the use of long system prompts and the accumulation of multi-turn conversational history needed to maintain context. 

As the process enters the model training and evaluation stage, overall token usage increases substantially. On the input side, the Task Manager continues to consume more tokens than the Model Manager and Model Developer. 
This is because it must retain full conversational memory to ensure coherent task tracking and a smooth user experience across iterations. For output tokens, the Model Manager becomes the dominant contributor. This is attributed to its two-step design of thinking and action: the thinking phase alone produces extensive textual output, making its total output volume the highest among all agents. The Model Developer follows, primarily due to the large number of tool function calls it issues.

Finally, during the deployment stage, the token consumption pattern becomes similar to that of the preparation phase, with the Task Manager and Deployment Operator engaging in a limited number of exchanges. Token usage here is relatively moderate, driven mainly by final instructions and confirmation steps.

Table \ref{tab:token-total} summarizes the overall token consumption and estimated cost for each agent throughout the full forecasting pipeline. A total of 201,534 input tokens and 24,732 output tokens were used, resulting in an approximate cost of \$0.751 under OpenAI's GPT-4o pricing. The Task Manager is the largest contributor to token usage, accounting for 47.2\% of input tokens, 18.0\% of output tokens, and a total cost of \$0.282. The Model Manager and Model Developer rank second and third in total cost, respectively. Notably, the Model Manager's output token cost exceeds its input cost, as output tokens are priced at four times the rate of input tokens under the GPT-4o pricing scheme.

\begin{table}[]
\centering 
\begin{threeparttable}
\caption{Summary of total input/output token consumption and estimated cost for each agent across the full forecasting workflow}
\label{tab:token-total}
\begin{tabular}{cccc}
\toprule
\textbf{Agent} & \textbf{Input Tokens} & \textbf{Output Tokens} & \textbf{Cost(\$)} \\ \midrule
\textbf{Task Manager} & 95,183 (47.2\%) & 4,452 (18.0\%) & 0.282 \\
\textbf{Preparation Assistant} & 20,761 (10.3\%) & 655 (2.6\%) & 0.059 \\
\textbf{Model Manager} & 38,665 (19.2\%) & 14,267 (57.7\%) & 0.239 \\
\textbf{Model Developer} & 28,890 (14.4\%) & 4,200 (17.0\%) & 0.114 \\
\textbf{Deployment Operator} & 18,035 (8.9\%) & 1,158 (4.7\%) & 0.057 \\
\textbf{Total} & 201,534 (100\%) & 24,732 (100\%) & 0.751 \\ \bottomrule
\end{tabular}
\begin{tablenotes}[para,flushleft]
\footnotesize
\textbf{Note:} Token cost estimates are based on the GPT-4o pricing at the time the experiments were conducted, with a rate of \$2.50 per million input tokens and \$10.00 per million output tokens.

\end{tablenotes}
\end{threeparttable}

\end{table}

Overall, the cost remains moderate and affordable, especially considering that GPT-4o is among the most advanced models currently available. Nonetheless, there is clear room for further optimization. Nearly two thirds of the cost arises from input tokens, suggesting that future work should focus on improving system prompt efficiency and memory management mechanisms to reduce redundancy and token overhead in long conversations.

\section{Conclusions}
\label{conclusion}
With the aim of lowering technical barriers of load forecasting and integrating contextual and human knowledge to better align models with real-world conditions, this paper proposes an LLM-based multi-agent collaboration framework for interactive load forecasting.We define the load forecasting workflow as comprising three stages—preparation, model training and evaluation, and postprocessing—and design specialized agents tailored to the specific demands of each stage, enabling human interaction throughout the process. We conduct experiments on two real-world datasets and validate the necessity of incorporating human interaction into the forecasting process. The results show that when human users provide appropriate insights, the framework can achieve improved forecasting performance. In addition, we evaluate the cost of running the framework and find it to be well within a practical range. For future work, advanced memory management methods will be further investigated, such as summarization-based memory compression and multilevel memory modules. Furthermore, explainable AI (XAI) techniques can be integrated into the framework to provide users with clearer reasoning behind model outputs, further enhancing trust and transparency in the interaction process.

\bibliographystyle{IEEEtran}
\bibliography{ref}




\end{document}